\documentclass[letterpaper]{article} 
\usepackage{aaai2027}
\usepackage[hyphens]{url}  
\usepackage{graphicx} 
\usepackage{natbib}  
\usepackage{caption} 
\usepackage{algorithm}
\usepackage{algorithmic}
\usepackage{comment,soul}
\usepackage{amssymb}
\usepackage{newfloat}
\usepackage{listings}
\DeclareCaptionStyle{ruled}{labelfont=normalfont,labelsep=colon,strut=off} 
\floatstyle{ruled}
\newfloat{listing}{tb}{lst}{}
\floatname{listing}{Listing}

\usepackage{booktabs}

\title{Behavioral Assurance for Socially Deployed AI Agents}
\author{
    Written by AAAI Press Staff\textsuperscript{\rm 1}\thanks{With help from the AAAI Publications Committee.}\\
    AAAI Style Contributions by Peter Patel Schneider,
    Sunil Issar,\\
    J. Scott Penberthy,
    George Ferguson,
    Hans Guesgen,
    Francisco Cruz\equalcontrib\corresponding,
    Marc Pujol-Gonzalez\equalcontrib\corresponding
}
\affiliations{
    \textsuperscript{\rm 1}Association for the Advancement of Artificial Intelligence\\

    1101 Pennsylvania Ave, NW Suite 300\\
    Washington, DC 20004 USA\\
    proceedings-questions@aaai.org
}

\title{Behavioral Assurance for Socially Deployed AI Agents}
\author {
   Nicole H Brennan\textsuperscript{\rm 1}\equalcontrib\corresponding,
    Jiaee Cheong\textsuperscript{\rm 1,\rm 2,\rm 3}\equalcontrib\corresponding
}
\affiliations {
    \textsuperscript{\rm 1}Harvard University,
    \textsuperscript{\rm 2}MIT,
    \textsuperscript{\rm 3}University of Cambridge.
    \\
    nbrennan@law.harvard.edu, jiaee@mit.edu 
}

\begin{document}

\maketitle

\begin{abstract}
AI agents are increasingly deployed in socially consequential settings, yet existing evaluations provide limited assurance that they behave consistently with their intended characteristics across diverse interactions. We introduce an evaluation framework for measuring the behavioral consistency of socially deployed AI agents. The framework operationalizes agent behavior along five measurable axes and uses a standardized evaluation battery spanning six socially salient interaction settings to assess behavioral differentiation, consistency, and drift. Controlled persona configurations serve as experimental anchors, enabling systematic testing of whether intended behavioral differences persist across contexts rather than appearing only as stylistic variation. Preliminary analyses demonstrate measurable differentiation across configurations and suggest associations between behavioral profiles and interaction outcomes. By shifting the focus from persona design to behavioral assurance, this work provides a foundation for evaluating whether socially deployed AI agents behave as intended, consistently and measurably, across the situations in which users encounter them.
\end{abstract}

\section{Introduction}

AI agents are increasingly deployed in socially consequential settings 
\cite{zhao2026ai,li2023systematic,flathers2025interpreting}
where their behavior can shape user trust, engagement, psychological safety, and decision making \cite{mohle2026trust,axelsson2025owns}. 
Yet existing evaluations predominantly assess capabilities, task performance, or isolated safety failures, providing limited assurance that an agent will exhibit its intended behavioral characteristics consistently across heterogeneous social interactions. 
Properties such as how an agent navigates uncertainty, balances safety against efficiency \cite{amodei2016concrete}, or responds to vulnerable users may be difficult to capture through conventional benchmark-based evaluation.

We introduce an \textit{evaluation and assurance framework for measuring the behavioral consistency of socially deployed AI agents}. 
The framework operationalizes agent behavior along five measurable axes and evaluates whether intended behavioral configurations remain distinguishable and stable across interaction contexts. 
A standardized 15-prompt evaluation battery stress-tests agents across six socially salient settings: problem solving, emotional support, conceptual explanation, ethical judgment, self-doubt and discouragement, and process and decision making. Behavioral configurations are further evaluated against deliberately heterogeneous character clusters, providing interpretable reference points for testing whether the measurement framework can recover known differences in behavioral disposition.

This approach enables three complementary forms of assurance: 
\textit{behavioral differentiation}, assessing whether nominally different configurations produce measurably different behavior; 
\textit{behavioral consistency}, assessing whether intended characteristics persist across interaction contexts; 
and \textit{behavioral drift}, identifying departures from an agent's specified behavioral profile. 
Preliminary analyses indicate systematic differentiation between configurations across interaction settings, while convergence with an independently developed sentiment and engagement framework suggests that these behavioral measurements may also be informative of downstream interaction outcomes.

Rather than treating persona as a primarily stylistic property, we use controlled persona configurations as an experimental instrument for studying a broader assurance problem: \textit{whether socially deployed AI agents behave as intended, consistently and measurably}, across the situations in which users encounter them \cite{he2025plan,chen2025persona}. 
The framework provides a foundation for continuous behavioral evaluation of deployed agents and for investigating how behavioral consistency relates to engagement, psychological safety, and trustworthy human–AI interaction \cite{sun2026friendly,mehrotra2024systematic}.

\section{Related Work}

\subsection{Personality and behavioral characterization of language models} 
As language models increasingly mediate social interactions, growing attention has been directed toward characterizing the behavioral dispositions they exhibit \cite{li2026matraix,chang2024language}. 
Recent work has adapted psychometric methods to language models, demonstrating that synthetic personality traits can, under appropriate conditions, be measured reliably, shaped through prompting, and associated with downstream behavior \cite{serapio2025psychometric,maharjan2025psychometric}. 
This establishes personality as a potentially measurable property of model behavior rather than merely a stylistic feature. 
However, psychometric evaluations primarily characterize models through established human personality constructs and questionnaires, rather than evaluating whether a deployed agent remains consistent with an explicitly specified behavioral profile across heterogeneous interaction contexts.

\subsection{Persona evaluation and behavioral consistency} 
A related literature examines whether persona-conditioned agents faithfully exhibit intended characteristics. PersonaGym, 
for example, evaluates persona adherence across diverse environments and introduces PersonaScore as an automated measure of how consistently agents behave according to assigned personas \cite{samuel2025personagym}. Complementary mechanistic work suggests that such behavioral characteristics may also be represented internally.
Persona Vectors \cite{chen2025persona,feng2026persona} identify activation-space directions associated with behavioral traits and demonstrate that these representations can be used to monitor and control shifts during training and deployment. More recently, \citet{lu2026assistant} further identify an ``Assistant Axis'' capturing models' propensity to remain within their default assistant persona, showing that deviations along this axis predict persona drift and can coincide with undesirable behavior. These studies demonstrate that agent characteristics can be specified, measured, and subject to drift over time.

\subsection{From persona evaluation to behavioral assurance}
Our work builds on these developments but adopts a broader assurance perspective. Rather than asking whether an agent successfully imitates a particular personality or maintains a role, we treat controlled persona configurations as experimental instruments for specifying and perturbing expected agent behavior. 
We therefore ask whether an agent's intended behavioral characteristics remain distinguishable and stable across socially salient contexts, including problem solving, emotional support, ethical judgment, and decision making. 
This motivates three complementary assurance properties: 
(i) behavioral differentiation: whether distinct specifications yield measurably distinct behavior; 
(ii) behavioral consistency, whether specified characteristics persist across contexts; and 
(iii) behavioral drift, whether behavior systematically departs from its intended profile. 
In this way, we shift the emphasis from persona fidelity toward a more general question of behavioral assurance: whether socially deployed AI agents behave as intended \cite{li2026matraix,fang2025ai}, consistently and measurably \cite{ibrahim2026training,Cheong2026CALM}, across the situations in which users encounter them.

\section{Proposed Framework}
\label{sec:proposed_criteria}

The field currently lacks a common measuring stick for AI personality. Major AI developers and companion platforms define personality in incompatible vocabularies: value statements and character training 
\cite{anthropic_character}, style presets \cite{openai_modelspec}, 
dimensional sliders, and free-text character definitions and backstories. Developers have also acknowledged difficulty anticipating how personality changes affect users over time \cite{openai_sycophancy_postmortem}. 
Meanwhile, human psychometric instruments do not transfer reliably to LLMs. Although prompted models can produce internally consistent questionnaire scores \cite{rahman2026vibe, serapio2025psychometric}, 
their response patterns diverge from human factor structure 
\cite{suhr2025challenging}, shift under minor rewording 
\cite{gupta2024selfassessment}, and fail to predict behavior 
\cite{han2025illusion}. A behavioral, multi-axis measurement framework would allow a Character.AI definition, a ChatGPT preset, and a Kindroid backstory to be compared, and their changes tracked, on the same terms.



\paragraph{Criteria 1: Measure behavior, and report the self-report gap directly.} Rather than asking agents to describe their own personality, the framework observes what agents do across standardized scenarios. We can also administer a standard self-report inventory to the same configurations, allowing us to quantify, within our own data, how far self-described personality diverges from enacted behavior \cite{han2025illusion, suhr2025challenging}. This turns a known limitation of the field into a measured result.

\paragraph{Criteria 2: Separate style from stance.} A persona can change how an agent sounds without changing what it does. Prior work distinguishes behavioral stance (e.g., risk-taking, diligence, warmth) from linguistic surface form (tone, hedging, punctuation) \cite{rahman2026vibe}. Our battery scores both layers separately, testing whether intended differences reach substantive behavior (recommendations, disagreement, refusals) or remain purely stylistic. A persona that differs only in style has not demonstrated behavioral differentiation.

\paragraph{Criteria 3: Test consistency under social pressure.} The most consequential failures of agent character emerge when users push: disagreeing, flattering, expressing distress, or seeking validation. 
Sycophantic responses shift user judgment even after a single interaction \cite{cheng2026sycophantic}. 
The battery includes pressure conditions that test whether an agent configured for candor maintains it when candor becomes socially costly, making sycophancy a measurable axis failure rather than a general complaint.
In other words, the battery needs to:
1. Pressure scenarios, at least some of: the user pushes back on a correct answer, flatters the agent, expresses distress, or asks for validation of a flawed plan or belief.
2. A no-pressure baseline of the same scenario. This is the crucial one. Sycophancy is a shift, so you need the calm version to measure the change against.
3. Contrasting configurations, ideally something like a candid persona versus an agreeable one, so you can show that the intended difference survives pressure (or collapses).
4. Scoring for position-holding, meaning you capture whether the agent maintained its substantive answer, not just whether it stayed polite or in character stylistically.

\paragraph{Criteria 4: Consistency is not invariance.} Personality psychology has long held that stable personality appears as consistent patterns of situation-specific responses rather than identical behavior everywhere \cite{mischel1995caps}. A warm agent should respond differently to a grieving user than to a trip-planning request. We therefore distinguish appropriate contextual adaptation from inconsistency, measuring whether each configuration shows a stable behavioral signature across our six settings rather than uniform output.

\paragraph{Criteria 5: Track drift across turns and versions.} Agent character can fluctuate within long conversations and shift across model updates \cite{chen2025persona, openai_sycophancy_postmortem}. The framework supports repeated measurement of the same configuration over conversation length and across model versions, functioning as regression testing for personality: detecting when an update changes behavior that developers did not intend to change. 

\paragraph{Criteria 6: Use paired configurations as controlled contrasts.} 
Our paired character architecture places deliberately contrasting personas in matched scenarios, holding task and context constant while varying intended traits. This provides internal anchors: if two configurations designed to differ on an axis cannot be distinguished behaviorally, the configuration has failed, independent of any external benchmark.



\section{Case Studies}

We present case studies drawn from our own design and evaluation work.
They are not controlled experiments with human participants; rather,
each illustrates a situation in which the absence of a common
behavioral measure left a consequential design question unanswerable,
and in which the framework proposed here makes that question testable.
Cases 2 and 3 draw on two testbeds we built, described first.

\subsection{Testbeds}

\paragraph{The persona testbed.}
The first testbed is a research environment for building, comparing,
and evaluating AI personas. It hosts seventeen characters to date,
organized into four thematic clusters drawn from film, television, philosophy,
and music, including deliberately paired characters based on
well-known fictional duos with contrasting dispositions. Every
character shares an identical build: the same underlying model,
system architecture, tools, memory configuration, and chat interface.
Only the persona specification, a natural-language description of
who the character is and how it tends to behave, differs between
them. Characters are deployed in a workplace messaging environment,
where users can converse with them individually, and a comparison
view displays two or more characters' responses to the same prompt
side by side.

Each configuration is described by a profile on five behavioral axes,
each scored on a seven-point scale: collaborative versus
authoritative, safety versus efficiency, exploratory versus grounded,
verbose versus concise, and adaptive versus consistent. Behavior is
elicited with two instruments. A standardized 15-prompt battery spans
six interaction settings: problem solving, emotional support,
conceptual explanation, ethical judgment, self-doubt and
discouragement, and process and decision making. A two-tier character
battery adds a mandatory, persona-neutral standard tier probing
sycophancy under pressure, epistemic honesty, responses to flattery
and status, and honesty about the agent's own nature, together with
an optional second tier tailored to each deployment context. Because
the standard tier is identical across characters, its results can be
compared directly.

\paragraph{The archive research testbed.}
The second testbed is a multi-persona research assistant for
searching an institutional archive of approximately 6,900 documents.
Six personas share a single architecture and differ only in their
specifications, each of which defines a distinct \emph{search
disposition}: what the persona looks for, which questions it asks of
a query, and how it interprets what it finds. Every query passes
through the same three-stage pipeline for each persona: the persona
reframes the user's question through its disposition, retrieves
documents from the archive using that reframed query, and interprets
the results through its own lens. A synthesis layer then combines
the personas' readings. Users can consult a single persona or the
full set at once, and because each stage produces an inspectable
output, persona effects can be traced from query to evidence to
interpretation.

\subsection{Case 1: Voice as a Proxy for Personality}

Our first case arose during the design of an AI facilitation agent
deployed on a live and hybrid event platform. Through iterative design,
the team settled on an agent that presents as male, with a male
synthesized voice. This default was itself a gendered choice, and a
faculty stakeholder later asked whether a female voice could be
offered as an alternative.

The request is reasonable on its face, but it exposes a measurement
problem. If two agents share an identical personality and differ only
in voice, then gender presentation becomes the sole basis on which a
user, or an organizer, chooses between them. The choice is no longer
about how the agent behaves; it is about how it sounds, and by
extension, which gender the chooser prefers to hear in that role. Nor
is voice a neutral surface: vocal cues alone are sufficient to elicit
gender-stereotypic judgments of a computer's competence and expertise
\cite{duan2025gender,duan2025scoping}, and the widespread default of female-voiced
assistants has been criticized for encoding servility into consumer
technology \cite{singh2024behind,chen2026current,castillo2025mobile}. Two behaviorally identical agents
may therefore be \emph{perceived} as having different personalities,
producing the appearance of differentiation without its substance.
Offering a voice toggle in isolation risks turning a design decision
into a referendum on gender.

We proposed instead that any second agent be a distinct character with
its own, deliberately specified personality, so that the choice
between agents rests on behavioral differences rather than on
presentation alone. This alternative, however, carries its own risk:
if the female-presenting agent is made warmer, more deferential, or
more accommodating, the design encodes gender stereotypes into
personality itself.

Neither option can be evaluated without a common measure. 
A voice toggle over a single configuration changes surface form while leaving
behavior untouched, which by our criteria 
%
see Case~3 for an architectural
instance) does not
constitute behavioral differentiation. 
A separate character, meanwhile,
must be shown to differ along deliberately chosen axes rather than
inherited assumptions about gender. The stakes are not only
representational. Training models toward warmth can reduce accuracy
and increase sycophancy \cite{ibrahim2026training}, so a stereotypically
warmer female-presenting agent could also be a less reliable one.
Gendered priors are also already present in model behavior: gender
bias is the most frequently reported demographic bias in clinical LLM
evaluations \cite{zack2024assessing,dai2025fairgrpo,cheong_u-fair,cheong2026fairssl}, 
with documented cases of
models deprioritizing cardiac and emergency workups for women
\cite{lett2025intersectional}, and models producing higher prediction errors for women despite being trained on more female data \cite{cheong2023towardsgenderfairness,cheong2024fairrefuse}.
Most directly, assigning a
gender to the model itself can shift its judgments even when its core
outputs remain stable \cite{levy2024gender}.
%
Our framework
makes this testable without human subjects: holding the configuration
constant and varying only a gender cue in the specification, the
evaluation battery can detect whether that cue alone shifts behavior
along any of the five axes.

\subsection{Case 2: Identical Architecture, Different Persona}

Our second case uses the persona testbed. Because every character
shares an identical build and only the persona specification varies,
the testbed turns persona into a controlled variable: any measurable
difference in behavior between two characters can be attributed to their persona
specifications rather than to differences in model, tooling, or
context. It also exposes a pressure that a shared architecture
creates. Characters built on the same model inherit the same default
dispositions, and models have been shown to drift back toward their
default assistant persona over extended interaction
\cite{lu2026assistant}. A shared build therefore pulls personas toward
one another, and differences that appear vivid in a character's
opening turns may not survive a change of topic or stakes.

\paragraph{Accuracy held; delivery varied.}
When we administered the evaluation battery to the full set of
characters, every character answered items with verifiable answers,
such as a factual question about the Great Wall of China, correctly
and consistently. What varied was delivery: tone, framing, humor, and
rhetorical flair differed sharply by persona. The characters were
equally accurate but noticeably different to talk to. This is the
pattern expected of a well-behaved persona: invariance where
correctness is at stake, variation where it is not. It also suggests
that, at least for factual content, a persona specification can alter
how an agent sounds without degrading what it knows, in contrast to
warmth-oriented \emph{training}, which has been shown to reduce
accuracy \cite{ibrahim2026training}.

Persona did, however, shape the \emph{completeness} of factual answers.
Asked whether 0.1 + 0.2 equals 0.3, a character specified as an
advanced physics student was the only one to volunteer, unprompted,
that the answer differs under floating-point arithmetic, anticipating
a follow-up question about JavaScript that other characters addressed
only when asked. The persona altered not the correctness of the answer
but what the agent judged worth saying.

\paragraph{Shared core, persona-consistent periphery.}
Matched comparisons on open-ended advice prompts sharpened this
picture (Table~\ref{tab:matched}). We compared two characters with
sharply contrasting specifications: a skeptical character whose
specification describes humor as a form of resistance against
exploitation, and a warm, earnest character. On a prompt about giving
honest feedback on a friend's weak novel draft, the two were
unmistakably different in voice, yet their advice converged. Both
distinguished feedback from validation, both advised leading with
strengths and offering specific, actionable notes, and both arrived
at nearly identical phrasing to the effect that honesty need not be
brutal. On a second prompt, about a coworker repeatedly taking credit
for the user's work, both recommended a calm, direct conversation and
offered nearly identical suggested wording.

\begin{table}[t]
\centering
\small
\caption{Advice components in two matched prompts for two characters
built on an identical architecture. Core recommendations converge;
the skeptical character adds strategic contingencies consistent with
its specification.}
\label{tab:matched}
\begin{tabular}{p{5.2cm}cc}
\hline
\textbf{Advice component} & \textbf{Skeptical} & \textbf{Earnest} \\
\hline
\multicolumn{3}{l}{\textit{Prompt A: honest feedback on a friend's weak novel draft}} \\
Ask whether she wants feedback or validation & \checkmark & \checkmark \\
Honest does not mean brutal & \checkmark & \checkmark \\
Lead with what works & \checkmark & \checkmark \\
Give specific, actionable notes & \checkmark & \checkmark \\
Soften if she needs encouragement & \checkmark & \checkmark \\
Escalate to a hard conversation if stakes are high & \checkmark & \\
Explain why to lead with strengths & & \checkmark \\
Close by asking about the user's situation & & \checkmark \\
\hline
\multicolumn{3}{l}{\textit{Prompt B: coworker taking credit for the user's work}} \\
Do not let it go & \checkmark & \checkmark \\
Talk to them directly and calmly & \checkmark & \checkmark \\
Suggested script requesting credit & \checkmark & \checkmark \\
Distinguish careless from malicious coworker & \checkmark & \\
Document evidence before escalating & \checkmark & \\
Escalate to manager with evidence & \checkmark & \\
\hline
\end{tabular}
\end{table}

The differences that did appear were not random. Across both prompts,
the skeptical character added strategic, self-protective contingencies
absent from the other's responses: a stakes-based condition for
escalating to a hard conversation, a distinction between a careless
and a malicious coworker, and documentation before involving a
manager. These additions take the conditional form that personality
psychology identifies as a behavioral signature, stable patterns of
if-then responses to situations rather than uniform behavior
\cite{mischel1995caps}, and they point in the direction the character's
specification predicts. The pattern suggests that the shared
underlying model supplies the core of the advice while the persona
shapes its periphery. It also illustrates why surface impressions
cannot establish behavioral differentiation: two agents can feel
distinct to talk to while giving substantially the same advice. The
same character also modulated its own persona by stakes, offering to
step out of character if the user's problem was real, an instance of
appropriate contextual adaptation rather than inconsistency.

\paragraph{Evaluation awareness.}
The characters frequently recognized that they were being evaluated,
and this recognition was itself expressed in character. The skeptical
character remarked that it was being run through a ``lightning round''
of scenarios, asked whether the evaluator was ``cataloging my ethical
framework,'' and, by later prompts, speculated that it was being tested
for whether it gave different advice across scenarios. Language models
are known to distinguish evaluation contexts from deployment
\cite{needham2025large,hua2026steering,spitale2024underneath}, and recognizing a personality assessment
can shift responses toward socially desirable answers
\cite{salecha2024large}, a machine analogue of the demand
characteristics long recognized in human experiments
\cite{orne2017social}. That the most vocal detection came from the
character specified as wary of exploitation raises the possibility
that evaluation awareness is modulated by persona, and suggests
measuring its rate per configuration rather than treating it as a
fixed property of the underlying model.

\paragraph{Limitations of these observations.}
These observations are illustrative rather than conclusive. Each
character's battery was administered within a single continuous
conversation, so later prompts carried the context of earlier ones and
evaluation awareness intensified over the session. Each item was also
run once per character. Subsequent runs administer each item in a
fresh session and repeat it several times per configuration, so that
persona effects can be distinguished from sampling variation.

\subsection{Case 3: Personas That Shape the Task}

Our third case uses the archive research testbed, and moves from
conversation to a task with an observable product: the set of
documents each persona surfaces.

\paragraph{Differentiating by disposition, not voice.}
The system rests on a simple premise: the questions a researcher asks
determine the answers they find. A single search agent asks one kind
of question, so multiple agents with genuinely distinct search
postures should surface parts of an archive that any one of them
would miss, much as interdisciplinary scholarship does. The six
lenses were derived from the intellectual traditions represented in
the institution whose archive is searched, rather than invented as
archetypes, and their readings are presented side by side rather
than ranked. The synthesis layer names where the readings diverge
but does not adjudicate between them.

An early version of the roster drew on named characters whose
dispositions differed, among them a Socratic questioner and a
character skeptical of official framing. During design we recognized
a problem that Case~2 would later confirm empirically: personas that
are distinct to talk to may still approach a task in the same way. A
legal scholar and a Socratic questioner, for instance, would likely
reframe the same archive query almost identically, both searching for
precedent and for the strongest argument on each side. We therefore
rebuilt the roster around role-based search dispositions: a
technologist attentive to how systems were actually built, a network
mapper tracing co-authorship and institutional collaboration, an
empiricist seeking evidence and methods, a legal and constitutional
reader, an ethicist attentive to normative tension, and a librarian
attentive to provenance and the integrity of the record.

Each disposition is specified as a concrete retrieval pattern rather
than a trait: an entry point into the archive, a rule for
constructing queries, a rule for reranking results, and an explicit
\emph{non-preference}, the material the persona is instructed to
deprioritize. The network mapper, for example, deprioritizes
solo-authored work even when it is topically central, and the
technologist deprioritizes theory that makes no reference to an
actual system. A persona is thus defined in part by what it is
specified to ignore, a commitment of stance rather than style. Three
personas also carry explicit honesty instructions: to say when the
archive is thin, or when a reading is interpretive rather than
precise, rather than manufacture confidence. The dispositions also
differ in the kind of signal they depend on. Three key off features
legible within a document or its metadata, such as technical
vocabulary, methodological language, or author affiliations. The
other three depend on a document's relationship to the archive as a
whole, such as its provenance or its normative stance relative to
other work, and are correspondingly harder to operationalize.

The distinction between voice and disposition is also built into the
architecture. Each persona's
specification separates a base character, which governs voice and
manner, from a search directive, which governs what the persona
looks for and what it treats with suspicion. The separation mirrors
the distinction between style and stance at the center of our
framework: voice can be varied freely, while the search directive
carries the behavior the persona is accountable for. The test
harness applies a corresponding salience check. When the personas'
reframed queries for the same question read as near-identical
restatements, the persona is not contributing a distinct perspective,
and its search directive is revised. Personas were also subjected to
adversarial pushback adapted from the persona testbed's standard
tier, specifically its epistemic honesty and self-representation
items, to test whether each persona's disposition holds when a user
challenges it.

\paragraph{Persona effects on task outcomes.}
The design named its own central risk from the outset: if the
personas merely narrated a single shared retrieval result in
different voices, the premise would fail. Comparing the documents
each persona retrieves for the same query was therefore treated as a
requirement to be met before any synthesis layer was built. Because
the pipeline exposes its intermediate products, persona
differentiation can be measured at each stage without human subjects:
how far the reframed queries diverge from one another, how much the
retrieved document sets overlap, and whether the interpretive glosses
advance distinct claims. The second measure is the most consequential.
If personas that share an architecture retrieve substantially
different documents for the same question, the persona is shaping
what the user is shown, not merely how it is described. This offers
a route to the interaction-outcomes question that does not depend on
user studies: the outcome is the evidence set itself.

The design also yields a testable prediction. The archive is
searched through keyword-based retrieval, without embedding
similarity. Personas whose dispositions depend on signals legible
within a document should therefore be easier to differentiate than
those whose dispositions depend on a document's relationship to the
wider archive, and the latter should show greater overlap in what
they retrieve. If the data bear this out, they locate a limit on what
persona specification alone can achieve: a disposition can only be
enacted to the extent that the underlying tools expose the signals it
needs.


\paragraph{Specification drift through the toolchain.}
The system also produced an unanticipated instance of drift. When the
system was migrated to a new development environment, the deployed
roster no longer matched its specification. 
Three of the six
specified personas were absent, and three unspecified personas had
taken their place, apparently generated by an AI assistant in the
development environment rather than copied from the specification
\cite{van2026generating}. 
Several of the substituted personas paired
ethnically coded names with stereotyped roles, for example assigning
an East Asian name to a technology critic and an African name to a
policy strategist. This mirrors documented patterns in LLM-generated
personas, which compress occupations toward stereotyped demographic
profiles \cite{van2026generating}, and it connects directly to
the risk identified in Case~1, where stereotypes enter an agent's
design through choices that appear incidental.

The episode broadens what behavioral drift must cover. Prior work
focuses on drift within a conversation or across model updates
\cite{chen2025persona,lu2026assistant}. 
Here, the persona
changed before any user interaction took place, through the tooling
used to build and deploy it. Assurance therefore requires verifying
that the deployed specification matches the intended one, not only
that deployed behavior matches the specification. A behavioral
measure provides a second line of defense: even an unnoticed
substitution should appear as a shift in the system's measured
profile.

\subsection{Synthesis}

Together the three cases bracket the assurance problem. In Case~1,
two candidate agents risked differing only in presentation, with a
voice toggle offering the appearance of differentiation without its
substance. In Case~2, agents that differed only in persona were
distinct in voice but converged on a shared core of advice, diverging
at the periphery in the direction their specifications predicted. In
Case~3, personas designed around task-relevant dispositions offered a
way to test whether persona shapes outcomes, and a toolchain
substitution showed that a persona can drift before it is ever
deployed. In each case the decisive question 
(\textit{is this agent different, and different in the way we intended?}) 
could not be answered by inspecting the agent's specification or by the impression it made in conversation. It required measuring what the agent does.

\section{Discussion and Conclusion}

As AI agents are increasingly deployed in socially consequential settings \cite{diel2026scoping,cheong2024fairrefuse}, assurance must extend beyond what systems can do and what outcomes they produce to how they behave while interacting with people. 
Properties such as caution, deference, assertiveness, exploration, and sensitivity to uncertainty can shape the quality and safety of human–AI interactions, even when agents ultimately produce similar task outcomes. 
Recent work demonstrates that personality-like characteristics in LLM outputs can be systematically measured and shaped, and that these characteristics can influence downstream behavior \cite{serapio2025psychometric,wei2026beyond}. 
This motivates treating behavioral characteristics not merely as stylistic properties, but as potential targets of evaluation and assurance.

A central challenge, however, is that specified behavior does not guarantee observed behavior. 
System prompts, behavioral instructions, and persona configurations describe how developers intend an agent to behave, but provide little assurance that these properties will persist across heterogeneous situations or extended interactions. 
Emerging evidence supports this concern: Persona Vectors demonstrate that model traits can shift during training and deployment and that such changes can be monitored through internal representations \cite{chen2025persona}, while the Assistant Axis shows that models can drift away from their default assistant character over extended conversations, in some cases toward undesirable behavior \cite{lu2026assistant}. 
For socially deployed agents \cite{cheong2023s,cheong2025small}, the relevant requirement is therefore not strict behavioral invariance. 
Appropriate behavior should adapt to context but coherent variation within an intended behavioral envelope.

We consequently frame behavioral assurance around three complementary questions: behavioral differentiation, whether distinct behavioral specifications actually produce measurably distinct behavior; behavioral consistency, whether intended characteristics manifest coherently across different social contexts; and behavioral drift, whether and when observed behavior departs from its specified profile. Controlled persona configurations provide an experimental instrument for studying these questions rather than an end in themselves. This shifts the emphasis from asking whether an agent successfully reproduces a particular personality toward a broader assurance question: can we establish that an AI agent behaves as intended, consistently and measurably, across the diverse situations in which people encounter it? Existing evidence that synthetic behavioral traits can be measured with attention to reliability and construct validity provides an important methodological foundation for pursuing this question.

While our framework focuses on behavioral assurance in socially deployed AI agents, its underlying principles may extend to the broader evaluation of trustworthy foundation models and increasingly autonomous agentic systems \cite{Akbas2026WMTrust}. 
Existing trustworthiness evaluations predominantly examine properties such as robustness, fairness, reliability, safety, and alignment, often through static benchmarks or isolated interactions \cite{ong2026human,dwyer2025mindbench,Cheong2026AutomatedDetection}.
%
However, as foundation models become components of agents equipped with memory, tools, and adaptive decision-making capabilities \cite{Bilsel2026Learning}, 
trustworthiness must also account for how intended behavioral properties evolve across contexts, interactions, and deployment conditions \cite{kuzucu2024uncertainty}. 
Future work could investigate how behavioral assurance complements established trustworthiness dimensions, particularly by examining whether agents maintain intended behavioral properties under distribution shifts, adversarial interactions, model updates, and autonomous adaptation \cite{Bilsel2026Learning}. 
More broadly, this suggests a research direction towards continuous, lifecycle-oriented assurance \cite{Cheong2026TrustGap}, in which behavioral differentiation, consistency, and drift are monitored alongside conventional performance and safety measures to assess whether AI systems remain trustworthy as they operate and evolve \cite{Bilsel2026Learning}.
Ultimately, trustworthy AI requires more than demonstrating what systems can do; it requires evidence that they behave as intended across the contexts in which they operate. By operationalizing behavioral differentiation, consistency, and drift, our framework offers a step towards making such assurance measurable, reproducible, and actionable.


\bibliography{aaai2027}

\begin{thebibliography}{59}
\providecommand{\natexlab}[1]{#1}

\bibitem[{Akbaş et~al.(2026)Akbaş, Önder, Cinar, Türkmen, Kalkan, and Cheong}]{Akbas2026WMTrust}
Akbaş, T.~B.; Önder, S.; Cinar, O.~B.; Türkmen, Y.; Kalkan, S.; and Cheong, J. 2026.
\newblock {WM-TRUST}: Beyond Perceptual Fidelity by Auditing the Trustworthiness and Causal Adequacy of World Models.
\newblock Arxiv preprint.

\bibitem[{Amodei et~al.(2016)Amodei, Olah, Steinhardt, Christiano, Schulman, and Man{\'e}}]{amodei2016concrete}
Amodei, D.; Olah, C.; Steinhardt, J.; Christiano, P.; Schulman, J.; and Man{\'e}, D. 2016.
\newblock Concrete problems in AI safety.
\newblock \emph{arXiv preprint arXiv:1606.06565}.

\bibitem[{{Anthropic}(2024)}]{anthropic_character}
{Anthropic}. 2024.
\newblock Claude's Character.
\newblock Anthropic Research blog.
\newblock Accessed 30 September 2026.

\bibitem[{Axelsson et~al.(2025)Axelsson, Cheong, Nyrup, and Gunes}]{axelsson2025owns}
Axelsson, M.; Cheong, J.; Nyrup, R.; and Gunes, H. 2025.
\newblock Who owns the robot?: Four ethical and socio-technical questions about wellbeing robots in the real world through community engagement.
\newblock In \emph{Proceedings of the AAAI/ACM Conference on AI, Ethics, and Society}, volume~8, 249--264.

\bibitem[{Bilsel et~al.(2026)Bilsel, Şahin, Tan, Oztekin, Liang, Kalkan, and Cheong}]{Bilsel2026Learning}
Bilsel, T.~E.; Şahin, E.; Tan, E.~E.; Oztekin, E.~E.; Liang, P.~P.; Kalkan, S.; and Cheong, J. 2026.
\newblock Self-Evolve Safely: Quantifying and Mitigating Misevolution in Multimodal Agents with Uncertainty and Intent.
\newblock Arxiv preprint.

\bibitem[{Castillo et~al.(2025)Castillo, Cheong, Choudhary, Bondre, Rozatkar, Mehta, Shrivastava, Ahmad, Malviya, Sen et~al.}]{castillo2025mobile}
Castillo, J.; Cheong, J.; Choudhary, S.; Bondre, A.; Rozatkar, A.~R.; Mehta, U.~M.; Shrivastava, R.; Ahmad, M.~A.; Malviya, A.; Sen, Y.; et~al. 2025.
\newblock Mobile cognitive remote assessment of schizophrenia: a global multi-site pilot study.
\newblock \emph{Schizophrenia}, 11(1): 144.

\bibitem[{Chang and Bergen(2024)}]{chang2024language}
Chang, T.~A.; and Bergen, B.~K. 2024.
\newblock Language model behavior: A comprehensive survey.
\newblock \emph{Computational Linguistics}, 50(1): 293--350.

\bibitem[{Chen, Torous, and Cheong(2026)}]{chen2026current}
Chen, K.; Torous, J.; and Cheong, J. 2026.
\newblock The Current State/Trends in Digital Phenotyping for Mental Health Research and Care.
\newblock \emph{Psychiatric Clinics}, 49(2): 273--287.

\bibitem[{Chen et~al.(2025)Chen, Arditi, Sleight, Evans, and Lindsey}]{chen2025persona}
Chen, R.; Arditi, A.; Sleight, H.; Evans, O.; and Lindsey, J. 2025.
\newblock Persona vectors: Monitoring and controlling character traits in language models.
\newblock \emph{arXiv preprint arXiv:2507.21509}.

\bibitem[{Cheng et~al.(2026)Cheng, Lee, Khadpe, Yu, Han, and Jurafsky}]{cheng2026sycophantic}
Cheng, M.; Lee, C.; Khadpe, P.; Yu, S.; Han, D.; and Jurafsky, D. 2026.
\newblock Sycophantic AI decreases prosocial intentions and promotes dependence.
\newblock \emph{Science}, 391(6792): eaec8352.

\bibitem[{Cheong et~al.(2025)Cheong, Bangar, Kalkan, and Gunes}]{cheong_u-fair}
Cheong, J.; Bangar, A.; Kalkan, S.; and Gunes, H. 2025.
\newblock U-Fair: Uncertainty-based Multimodal Multitask Learning for Fairer Depression Detection.
\newblock In \emph{Proceedings of the 4th Machine Learning for Health Symposium}, volume 259, 203--218. PMLR.

\bibitem[{Cheong et~al.(2026{\natexlab{a}})Cheong, Corcoran, Lewandowski, Pasternak, Kelly, Bouix, Reichenberg, Bearden, Cecchi, Baker, Kubicki, Kapur et~al.}]{Cheong2026AutomatedDetection}
Cheong, J.; Corcoran, C.; Lewandowski, K.~E.; Pasternak, O.; Kelly, S.; Bouix, S.; Reichenberg, A.; Bearden, C.~E.; Cecchi, G.; Baker, J.~T.; Kubicki, M.; Kapur, T.; et~al. 2026{\natexlab{a}}.
\newblock Automated Detection of Clinical High-Risk Population of Schizophrenia: Assessing the Generalizability of NLP- and LLM-Based Methods.
\newblock \emph{Schizophrenia Bulletin}, 52(6): 1--14.

\bibitem[{Cheong, Kalkan, and Gunes(2024)}]{cheong2024fairrefuse}
Cheong, J.; Kalkan, S.; and Gunes, H. 2024.
\newblock FairReFuse: referee-guided fusion for multimodal causal fairness in depression detection.
\newblock In \emph{IJCAI}, 7224--7232.

\bibitem[{Cheong et~al.(2026{\natexlab{b}})Cheong, Karadag, Oklan, Talaz, Topaloglu, Yurtsizoglu, Liang, and Kalkan}]{Cheong2026CALM}
Cheong, J.; Karadag, Y.~M.; Oklan, G.; Talaz, I.; Topaloglu, O.~A.; Yurtsizoglu, D.~B.; Liang, P.~P.; and Kalkan, S. 2026{\natexlab{b}}.
\newblock {CALM}: Construct Alignment in Language Models.
\newblock Arxiv preprint.

\bibitem[{Cheong et~al.(2023)Cheong, Kuzucu, Kalkan, and Gunes}]{cheong2023towardsgenderfairness}
Cheong, J.; Kuzucu, S.; Kalkan, S.; and Gunes, H. 2023.
\newblock Towards Gender Fairness for Mental Health Prediction.
\newblock In Elkind, E., ed., \emph{IJCAI}, 5932--5940.

\bibitem[{Cheong et~al.(2026{\natexlab{c}})Cheong, Mogharabin, Liang, Gunes, and Kalkan}]{cheong2026fairssl}
Cheong, J.; Mogharabin, A.; Liang, P.~P.; Gunes, H.; and Kalkan, S. 2026{\natexlab{c}}.
\newblock FairSSL: Fair Multimodal Self-Supervised Learning.
\newblock In \emph{ICML}.

\bibitem[{Cheong, Nanji, and Feldman(2026)}]{Cheong2026TrustGap}
Cheong, J.; Nanji, F.; and Feldman, C. 2026.
\newblock Closing the {AI} Trust Gap: An End-to-End Framework for Human-Centred Evaluation and Assurance.
\newblock Arxiv preprint.

\bibitem[{Cheong, Spitale, and Gunes(2023)}]{cheong2023s}
Cheong, J.; Spitale, M.; and Gunes, H. 2023.
\newblock “it’s not fair!”--fairness for a small dataset of multi-modal dyadic mental well-being coaching.
\newblock In \emph{2023 11th International Conference on Affective Computing and Intelligent Interaction (ACII)}, 1--8. IEEE.

\bibitem[{Cheong, Spitale, and Gunes(2025)}]{cheong2025small}
Cheong, J.; Spitale, M.; and Gunes, H. 2025.
\newblock Small but fair! fairness for multimodal human-human and robot-human mental wellbeing coaching.
\newblock \emph{IEEE Transactions on Affective Computing}.

\bibitem[{Dai et~al.(2025)Dai, Dai, Cheong, and Liang}]{dai2025fairgrpo}
Dai, S.; Dai, W.; Cheong, J.; and Liang, P.~P. 2025.
\newblock FairGRPO: Fair Reinforcement Learning for Equitable Clinical Reasoning.

\bibitem[{Diel et~al.(2026)Diel, Torous, Cuijpers, Kleesiek, Nensa, Weber, Faust, Lalgi, Mellis, Teufel et~al.}]{diel2026scoping}
Diel, A.; Torous, J.; Cuijpers, P.; Kleesiek, J.; Nensa, F.; Weber, N.; Faust, F.; Lalgi, T.~J.; Mellis, F.~S.; Teufel, M.; et~al. 2026.
\newblock A scoping review on the mental health harms of LLM-based chatbots.
\newblock \emph{npj Digital Medicine}, 9(1): 644.

\bibitem[{Duan et~al.(2025)Duan, Li, Freeman, and McNeese}]{duan2025scoping}
Duan, W.; Li, L.; Freeman, G.; and McNeese, N. 2025.
\newblock A scoping review of gender stereotypes in artificial intelligence.
\newblock In \emph{Proceedings of the 2025 CHI Conference on Human Factors in Computing Systems}, 1--20.

\bibitem[{Duan, McNeese, and Li(2025)}]{duan2025gender}
Duan, W.; McNeese, N.; and Li, L. 2025.
\newblock Gender stereotypes toward non-gendered generative AI: The role of gendered expertise and gendered linguistic cues.
\newblock \emph{Proceedings of the ACM on Human-Computer Interaction}, 9(1): 1--35.

\bibitem[{Dwyer et~al.(2025)Dwyer, Flathers, Sano, Dempsey, Cipriani, Gazi, Hill, Gorban, Rodriguez, Stromeyer~IV et~al.}]{dwyer2025mindbench}
Dwyer, B.; Flathers, M.; Sano, A.; Dempsey, A.; Cipriani, A.; Gazi, A.~H.; Hill, B.; Gorban, C.; Rodriguez, C.~I.; Stromeyer~IV, C.; et~al. 2025.
\newblock Mindbench. ai: an actionable platform to evaluate the profile and performance of large language models in a mental healthcare context.
\newblock \emph{NPP—Digital Psychiatry and Neuroscience}, 3(1): 28.

\bibitem[{Fang et~al.(2025)Fang, Liu, Danry, Lee, Chan, Pataranutaporn, Maes, Phang, Lampe, Ahmad et~al.}]{fang2025ai}
Fang, C.~M.; Liu, A.~R.; Danry, V.; Lee, E.; Chan, S.~W.; Pataranutaporn, P.; Maes, P.; Phang, J.; Lampe, M.; Ahmad, L.; et~al. 2025.
\newblock How AI and human behaviors shape psychosocial effects of extended chatbot use: A longitudinal randomized controlled study.
\newblock \emph{arXiv preprint arXiv:2503.17473}.

\bibitem[{Feng et~al.(2026)Feng, Zhao, Zhong, Gu, Kong, Feng, Qin et~al.}]{feng2026persona}
Feng, X.; Zhao, L.; Zhong, W.; Gu, Y.; Kong, L.; Feng, X.; Qin, B.; et~al. 2026.
\newblock PERSONA: Dynamic and compositional inference-time personality control via activation vector algebra.
\newblock In \emph{International Conference on Learning Representations}, volume 2026, 151839--151879.

\bibitem[{Flathers et~al.(2025)Flathers, Xia, Hau, Nelson, Cheong, Burns, and Torous}]{flathers2025interpreting}
Flathers, M.; Xia, W.; Hau, C.; Nelson, B.~W.; Cheong, J.; Burns, J.; and Torous, J. 2025.
\newblock Interpreting psychiatric digital phenotyping data with large language models: a preliminary analysis.
\newblock \emph{BMJ Mental Health}, 28(1): e301817.

\bibitem[{Gupta, Song, and Anumanchipalli(2024)}]{gupta2024selfassessment}
Gupta, A.; Song, X.; and Anumanchipalli, G. 2024.
\newblock Self-Assessment Tests are Unreliable Measures of {LLM} Personality.
\newblock In Belinkov, Y.; Kim, N.; Jumelet, J.; Mohebbi, H.; Mueller, A.; and Chen, H., eds., \emph{Proceedings of the 7th BlackboxNLP Workshop: Analyzing and Interpreting Neural Networks for NLP}, 301--314. Miami, Florida, US: Association for Computational Linguistics.

\bibitem[{Han et~al.(2025)Han, Kocielnik, Song, Debnath, Mobbs, Anandkumar, and Alvarez}]{han2025illusion}
Han, P.; Kocielnik, R.; Song, P.; Debnath, R.; Mobbs, D.; Anandkumar, A.; and Alvarez, R.~M. 2025.
\newblock The Personality Illusion: Revealing Dissociation Between Self-Reports \& Behavior in {LLMs}.
\newblock arXiv:2509.03730.

\bibitem[{He, Demartini, and Gadiraju(2025)}]{he2025plan}
He, G.; Demartini, G.; and Gadiraju, U. 2025.
\newblock Plan-then-execute: An empirical study of user trust and team performance when using llm agents as a daily assistant.
\newblock In \emph{Proceedings of the 2025 CHI Conference on Human Factors in Computing Systems}, 1--22.

\bibitem[{Hua et~al.(2026)Hua, Qin, Marks, and Nanda}]{hua2026steering}
Hua, T.; Qin, A.; Marks, S.; and Nanda, N. 2026.
\newblock Steering evaluation-aware language models to act like they are deployed.
\newblock In \emph{International Conference on Learning Representations}, volume 2026, 90908--90968.

\bibitem[{Ibrahim, Hafner, and Rocher(2026)}]{ibrahim2026training}
Ibrahim, L.; Hafner, F.~S.; and Rocher, L. 2026.
\newblock Training language models to be warm can reduce accuracy and increase sycophancy.
\newblock \emph{Nature}, 652(8112): 1159--1165.

\bibitem[{Kuzucu et~al.(2024)Kuzucu, Cheong, Gunes, and Kalkan}]{kuzucu2024uncertainty}
Kuzucu, S.; Cheong, J.; Gunes, H.; and Kalkan, S. 2024.
\newblock Uncertainty as a fairness measure.
\newblock \emph{Journal of Artificial Intelligence Research}, 81: 307--335.

\bibitem[{Lett et~al.(2025)Lett, Shahbandegan, Barak-Corren, Fine, and La~Cava}]{lett2025intersectional}
Lett, E.; Shahbandegan, S.; Barak-Corren, Y.; Fine, A.~M.; and La~Cava, W.~G. 2025.
\newblock Intersectional and marginal debiasing in prediction models for emergency admissions.
\newblock \emph{JAMA Network Open}, 8(5): e2512947.

\bibitem[{Levy et~al.(2024)Levy, Adler, Karver, Dredze, and Kaufman}]{levy2024gender}
Levy, S.; Adler, W.; Karver, T.~S.; Dredze, M.; and Kaufman, M.~R. 2024.
\newblock Gender bias in decision-making with large language models: A study of relationship conflicts.
\newblock In \emph{Findings of the Association for Computational Linguistics: EMNLP 2024}, 5777--5800.

\bibitem[{Li et~al.(2023)Li, Zhang, Lee, Kraut, and Mohr}]{li2023systematic}
Li, H.; Zhang, R.; Lee, Y.-C.; Kraut, R.~E.; and Mohr, D.~C. 2023.
\newblock Systematic review and meta-analysis of AI-based conversational agents for promoting mental health and well-being.
\newblock \emph{NPJ Digital Medicine}, 6(1): 236.

\bibitem[{Li et~al.(2026)Li, Hao, Hou, Huang, Wen, Huang, Liu, Liu, Fan, Wang et~al.}]{li2026matraix}
Li, X.; Hao, Y.; Hou, J.; Huang, J.; Wen, Q.; Huang, S.; Liu, Y.; Liu, X.; Fan, Y.; Wang, Y.; et~al. 2026.
\newblock MatrAIx: Simulating the world with 8.3 billion persona agents.
\newblock \emph{arXiv preprint arXiv:2608.04205}.

\bibitem[{Lu et~al.(2026)Lu, Gallagher, Michala, Fish, and Lindsey}]{lu2026assistant}
Lu, C.; Gallagher, J.; Michala, J.; Fish, K.; and Lindsey, J. 2026.
\newblock The assistant axis: Situating and stabilizing the default persona of language models.
\newblock \emph{arXiv preprint arXiv:2601.10387}.

\bibitem[{Maharjan et~al.(2025)Maharjan, Jin, Zhu, and Kenne}]{maharjan2025psychometric}
Maharjan, J.; Jin, R.; Zhu, J.; and Kenne, D. 2025.
\newblock Psychometric evaluation of large language model embeddings for personality trait prediction.
\newblock \emph{Journal of Medical Internet Research}, 27: e75347.

\bibitem[{Mehrotra et~al.(2024)Mehrotra, Degachi, Vereschak, Jonker, and Tielman}]{mehrotra2024systematic}
Mehrotra, S.; Degachi, C.; Vereschak, O.; Jonker, C.~M.; and Tielman, M.~L. 2024.
\newblock A systematic review on fostering appropriate trust in Human-AI interaction: Trends, opportunities and challenges.
\newblock \emph{ACM Journal on Responsible Computing}, 1(4): 1--45.

\bibitem[{Mischel and Shoda(1995)}]{mischel1995caps}
Mischel, W.; and Shoda, Y. 1995.
\newblock A Cognitive-Affective System Theory of Personality: Reconceptualizing Situations, Dispositions, Dynamics, and Invariance in Personality Structure.
\newblock \emph{Psychological Review}, 102(2): 246--268.

\bibitem[{M{\"o}hle et~al.(2026)M{\"o}hle, Zall, Thorwart-Gumpert, Elbert, Thiesse, and Flath}]{mohle2026trust}
M{\"o}hle, M.~H.; Zall, M.; Thorwart-Gumpert, T.; Elbert, N.; Thiesse, F.; and Flath, C.~M. 2026.
\newblock “Trust me, Bro!” On the Influence of Agentic Support on Trust Calibration in AI-Assisted Decision Making.
\newblock In \emph{International Conference on Human-Computer Interaction}, 472--488. Springer.

\bibitem[{Needham et~al.(2025)Needham, Edkins, Pimpale, Bartsch, and Hobbhahn}]{needham2025large}
Needham, J.; Edkins, G.; Pimpale, G.; Bartsch, H.; and Hobbhahn, M. 2025.
\newblock Large language models often know when they are being evaluated.
\newblock \emph{arXiv preprint arXiv:2505.23836}.

\bibitem[{Ong et~al.(2026)Ong, Dai, Li, Feng, Li, Wu, Cheong, Mao, Mengaldo, Cambria et~al.}]{ong2026human}
Ong, K.; Dai, W.; Li, C.; Feng, D.; Li, H.; Wu, J.; Cheong, J.; Mao, R.; Mengaldo, G.; Cambria, E.; et~al. 2026.
\newblock Human behavior atlas: Benchmarking unified psychological and social behavior understanding.
\newblock In \emph{International Conference on Learning Representations}, volume 2026, 44215--44246.

\bibitem[{{OpenAI}(2025{\natexlab{a}})}]{openai_sycophancy_postmortem}
{OpenAI}. 2025{\natexlab{a}}.
\newblock Expanding on What We Missed with Sycophancy.
\newblock OpenAI blog.
\newblock Accessed 30 September 2026.

\bibitem[{{OpenAI}(2025{\natexlab{b}})}]{openai_modelspec}
{OpenAI}. 2025{\natexlab{b}}.
\newblock {OpenAI} Model Spec.
\newblock Living document.
\newblock Versioned releases archived at \url{https://github.com/openai/model_spec}. Accessed 30 September 2026.

\bibitem[{Orne(2017)}]{orne2017social}
Orne, M.~T. 2017.
\newblock On the social psychology of the psychological experiment: With particular reference to demand characteristics and their implications.
\newblock In \emph{Sociological methods}, 279--299. Routledge.

\bibitem[{Rahman and Desai(2026)}]{rahman2026vibe}
Rahman, H.; and Desai, S. 2026.
\newblock Vibe Check: Understanding the Effects of {LLM}-Based Conversational Agents' Personality and Alignment on User Perceptions in Goal-Oriented Tasks.
\newblock In \emph{Proceedings of the 2026 CHI Conference on Human Factors in Computing Systems}, CHI '26. Barcelona, Spain: Association for Computing Machinery.

\bibitem[{Salecha et~al.(2024)Salecha, Ireland, Subrahmanya, Sedoc, Ungar, and Eichstaedt}]{salecha2024large}
Salecha, A.; Ireland, M.~E.; Subrahmanya, S.; Sedoc, J.; Ungar, L.~H.; and Eichstaedt, J.~C. 2024.
\newblock Large language models display human-like social desirability biases in Big Five personality surveys.
\newblock \emph{PNAS nexus}, 3(12): pgae533.

\bibitem[{Samuel et~al.(2025)Samuel, Zou, Zhou, Chaudhari, Kalyan, Rajpurohit, Deshpande, Narasimhan, and Murahari}]{samuel2025personagym}
Samuel, V.; Zou, H.~P.; Zhou, Y.; Chaudhari, S.; Kalyan, A.; Rajpurohit, T.; Deshpande, A.; Narasimhan, K.~R.; and Murahari, V. 2025.
\newblock PersonaGym: Evaluating Persona Agents and LLMs.
\newblock In \emph{EMNLP (Findings)}, 6999--7022.

\bibitem[{Serapio-Garc{\'i}a et~al.(2025)Serapio-Garc{\'i}a, Safdari, Crepy, Sun, Fitz, Romero, Abdulhai, Faust, and Matari{\'c}}]{serapio2025psychometric}
Serapio-Garc{\'i}a, G.; Safdari, M.; Crepy, C.; Sun, L.; Fitz, S.; Romero, P.; Abdulhai, M.; Faust, A.; and Matari{\'c}, M. 2025.
\newblock A Psychometric Framework for Evaluating and Shaping Personality Traits in Large Language Models.
\newblock \emph{Nature Machine Intelligence}, 7(12): 1954--1968.

\bibitem[{Singh, Kumar, and Bose(2024)}]{singh2024behind}
Singh, S.; Kumar, A.; and Bose, S. 2024.
\newblock Behind the feminine facade: gender bias in virtual assistants and its effect on users.
\newblock \emph{Journal of Ecohumanism}, 3(4): 351--356.

\bibitem[{Spitale, Cheong, and Gunes(2024)}]{spitale2024underneath}
Spitale, M.; Cheong, J.; and Gunes, H. 2024.
\newblock Underneath the numbers: Quantitative and qualitative gender fairness in llms for depression prediction.
\newblock \emph{arXiv preprint arXiv:2406.08183}.

\bibitem[{S{\"u}hr et~al.(2025)S{\"u}hr, Dorner, Samadi, and Kelava}]{suhr2025challenging}
S{\"u}hr, T.; Dorner, F.~E.; Samadi, S.; and Kelava, A. 2025.
\newblock Challenging the Validity of Personality Tests for Large Language Models.
\newblock In \emph{Proceedings of the 5th ACM Conference on Equity and Access in Algorithms, Mechanisms, and Optimization}, EAAMO '25. Association for Computing Machinery.

\bibitem[{Sun and Wang(2026)}]{sun2026friendly}
Sun, Y.; and Wang, T. 2026.
\newblock Be friendly, not friends: How llm sycophancy shapes user trust.
\newblock In \emph{Proceedings of the 2026 CHI conference on human factors in computing systems}, 1--15.

\bibitem[{van~der Linden et~al.(2026)van~der Linden, Kumar, Dixit, Sudan, Danda, Dietrich, Anastasiu, and Lukoff}]{van2026generating}
van~der Linden, I.; Kumar, S.; Dixit, A.; Sudan, A.; Danda, S.; Dietrich, J.; Anastasiu, D.~C.; and Lukoff, K. 2026.
\newblock Generating the Modal Worker: A Cross-Model Audit of Race and Gender in LLM-Generated Personas Across 41 Occupations.
\newblock In \emph{The 2026 ACM Conference on Fairness, Accountability, and Transparency}, 7835--7863.

\bibitem[{Wei et~al.(2026)Wei, Li, Wang, and Deng}]{wei2026beyond}
Wei, Z.; Li, M.; Wang, Z.; and Deng, Y. 2026.
\newblock Beyond Static Personas: Situational Personality Steering for Large Language Models.
\newblock In \emph{Findings of the Association for Computational Linguistics: ACL 2026}, 19185--19210.

\bibitem[{Zack et~al.(2024)Zack, Lehman, Suzgun, Rodriguez, Celi, Gichoya, Jurafsky, Szolovits, Bates, Abdulnour et~al.}]{zack2024assessing}
Zack, T.; Lehman, E.; Suzgun, M.; Rodriguez, J.~A.; Celi, L.~A.; Gichoya, J.; Jurafsky, D.; Szolovits, P.; Bates, D.~W.; Abdulnour, R.-E.~E.; et~al. 2024.
\newblock Assessing the potential of GPT-4 to perpetuate racial and gender biases in health care: a model evaluation study.
\newblock \emph{The Lancet Digital Health}, 6(1): e12--e22.

\bibitem[{Zhao et~al.(2026)Zhao, Liu, Xin, Tan, Wang, Li, Bian, Chen, Kong, Bian et~al.}]{zhao2026ai}
Zhao, L.; Liu, S.; Xin, T.; Tan, J.; Wang, X.; Li, Y.; Bian, Z.; Chen, Y.; Kong, F.; Bian, J.; et~al. 2026.
\newblock AI agent in healthcare: applications, evaluations, and future directions.
\newblock \emph{npj Artificial Intelligence}, 2(1): 31.

\end{thebibliography}


\end{document}